# A Study on the Behavior of a Neural Network for Grouping the Data

**Suneetha Chittineni [1] and Dr.Raveendra Babu Bhogapathi [2]**

[1]**Department of Computer Applications,
R.V.R. & J.C. College of Engineering,
Guntur, India.**
chittineni_es@yahoo.com

[2]**Director-Operations
DELTA Technology & Management Services Pvt. Ltd.,
Hyderabad, India.**
rbhogapathi@yahoo.com

**Abstract**
One of the frequently stated advantages of neural networks is that they can work effectively with non-normally distributed data. But optimal results are possible with normalized data.In this paper, how normality of the input affects the behaviour of a K-means fast learning artificial neural network(KFLANN) for grouping the data is presented. Basically, the grouping of high dimensional input data is controlled by additional neural network input parameters namely vigilance and tolerance. Neural networks learn faster and give better performance if the input variables are pre-processed before being fed to the input units of the neural network. A common way of dealing with data that is not normally distributed is to perform some form of mathematical transformation on the data that shifts it towards a normal distribution.In a neural network, data preprocessing transforms the data into a format that will be more easily and effectively processed for the purpose of the user. Among various methods, Normalization is one which organizes data for more efficient access. Experimental results on several artificial and synthetic data sets indicate that the groups formed in the data vary with non-normally distributed data and normalized data and also depends on the normalization method used.

## 1. Introduction

Data preprocessing describes any type of processing on atomic data that has not been processed further to preparing it for another processing procedure. Different data preprocessing techniques like cleaning method, outlier detection, data integration and transformation can be carried out before clustering process to achieve successful analysis. Normalization is an important preprocessing step in data mining to standardize the values of all variables from dynamic range into specific range.

Typical objective functions in clustering formalize the goal of attaining high intra-cluster similarity (Patterns within a cluster are similar) and low inter-cluster similarity (Patterns from different clusters are dissimilar). This is an internal criterion for the quality of a clustering.

In unsupervised learning, an output unit is trained to respond to clusters of pattern within the input. In this paradigm, the system is supposed to discover statistically salient features of the input population. Unlike the supervised learning paradigm, there is no a priori set of categories into which the patterns are to be classified; rather, the system must develop its own representation of the input stimuli.

Supervised networks like Simple Perceptrons, Back Propagation (BP), and Radial Basis Function (RBF) networks need a teacher to tell the network what the desired output should be. Unsupervised nets include Kohonen Self Organizing maps (SOM), Adaptive Resonance theory (ART) etc. The major applications of unsupervised nets include clustering data, reducing the dimensionality of the data. In data clustering, exactly one of small number of output units comes on in response to an input. In dimension reduction, large number of input units is compressed into a small number of output units.

Kohonen clustering algorithm takes high-dimensional input, clusters it, and retaining some topological ordering of the output [7]. Adaptive Resonance Theory (ART) was initially introduced by Grossberg (1976) as a theory of human information processing [3]. ART neural networks are extensively used for supervised, unsupervised classification tasks and functional approximation. There exist many different variations of ART networks today (Carpenter and Grossberg, 1998). ART1 performs unsupervised learning for binary input patterns, ART2 is modified to handle both analog and binary input patterns, and ART3 performs parallel searches of distributed recognition codes in a multilevel network hierarchy. Fuzzy ARTMAP represents a synthesis of elements from neural networks, expert systems, and fuzzy logic.

Many variations of fast learning artificial neural network algorithms have been proposed. A fast learning artificial neural network (FLANN) models was first developed by Tay and Evans [17] to solve a set of problems in the area of pattern classification. FLANN [3] [4] was designed with concepts found in ART but imposed the Winner Take All (WTA) property within the algorithm. Further improvement was done to take in numerical continuous value in FLANN II [11]. The original FLANN II was restricted by its sensitivity to the pattern sequence. This was later overcome by the inclusion of k-means calculations, which served to remove inconsistent cluster formations [13]. The KFLANN utilizes the Leader-type algorithm first addressed by Hartigan [9] and also draws some parallel similarities established in the Adaptive Resonance Theories developed by Grossberg [7] and later ART algorithms by Carpenter et al [2].The later improvement on KFLANN [15] includes data point reshuffling which resolves the data sequence sensitivity that creates stable clusters. Clusters are said to be stable if the cluster formation is complete after some iterations and the cluster centers remain consistent.

The organization of this paper is as follows. In section II, an overview of the data normalization methods are described. Section III gives the details of neural network input parameters. Section IV presents various formulas for computing tolerance values of the attributes. In Section V the K-means Fast Learning Artificial Neural Network algorithm is presented. Section VI presents the experimental analysis of the results and conclusions follow in section VII.

## 2. Data normalization

### 1) Z-score normalization

The *Z*-score is called as standardized unit. It indicates how far and in what direction a variable deviates from its distribution's mean, expressed in units of its distribution's standard deviation. The *Z*-scores are especially informative when the distribution to which they refer is normal. If the value of *Z* is positive, it means the variable *X* is above its mean and if Z is negative, then *X* is below its mean (Cryer & Miller, 1994). It can also measure the outliers of a dataset. If the value of a *Z*-score is greater than 3, it indicates that the data distribution has outliers (Tamhane & Dunlop, 2000). It can be defined as follows:

$$z-score = \frac{(x-\bar{x})}{\sigma} \quad (1)$$

### 2) Min-Max normalization

Min-max normalization performs a linear transformation on the original data. Suppose that $min_a$ and $max_a$ are the minimum and the maximum values for attribute A. Min-max normalization maps a value v of A to v' in the range [new-$min_a$,new-$max_a$] by computing

$$v' = \left(new\_max_a - new\_min_a\right) * \left(\frac{(v-min_a)}{(max_a - min_a)}\right) + new\_min_a \quad (2)$$

## 3. Neural Network Input Parameters

KFLANN is an atomic module, suitable for creating scalable networks which performs similar processing evident within biological neural systems. The original FLANN and KFLANN share the exact same network parameters. Both algorithms require an initialization of two network parameters, namely the Vigilance factor ($\rho$) and the Tolerance setting ($\delta$). These network parameters are the key to obtain an optimal clustering outcome.

### 3.1 Vigilance

The Vigilance ($\rho$) is a parameter that originated from [2]. It was designed as a means to influence the matching degree between the current exemplar and long term memory trace. The higher the $\rho$ value, the stricter the match, while for a smaller $\rho$ value, a more relaxed matching criteria is set. The $\rho$ value in the KFLANN is similar and it is used to determine the number of the attributes in the current exemplar that is similar to the selected output node. For example, if a pattern consists of 12 attributes and a clustering criteria was set such that a similarity of 4 attributes was needed for consideration into the same cluster, then the $\rho$ should be held at 3. The Vigilance formulation is given by equation (3)

$$\rho = \frac{f_{match}}{f_{total}} \quad (3)$$

If the vigilance value is high more number of clusters were formed than when it was set lower.

## 3.2 Tolerance

Tolerance parameter provides the localized control, affects individual input features where as the Global variation in the input features is governed by vigilance parameter. Tolerance setting, δ, is the measurement of a particular feature consistency that measures the maximum range that the specified feature is allowed to fluctuate. The tolerance setting can be performed through various methodologies. Three methods are described in the sections that follow. The tolerance setting of the exemplar attributes is the measurement of attributes dispersion, and thus computation is performed for every feature of the training exemplar at the initial stage.

1. The presence of Domain knowledge is perhaps the most helpful form of tolerance setting. With the existence of knowledge from the domain expert, the tolerance of pattern attributes may be determined. This approach is able to produce an acceptable clustering result since the domain knowledge is usually superior. However, domain knowledge is not always available. Two other approaches have been included to harness the clustering capabilities of K-FLANN even in the absence of domain knowledge.

2. In the mathematical context, Standard deviation is the measurement of dispersion
3. of a particular variable from the mean value [1]. Standard deviation method used as the tolerance setting of each feature. The tolerance computation formula is given in equation (4). From the equation, it indicates that it is susceptible to outlier's values, which may increase the δ value. This method is appropriate only if the variation of each feature in exemplar is uniform. There should be a minimal presence of the extreme points lying in the outer regions of the main cluster body. However in most pattern classification problems, the data sets are usually attach with the outliers points.

$$p\sigma_d = \partial_d = \sqrt{\frac{\sum_{k=1}^{n}(x_{dk} - \mu_d)^2}{n}} \quad (4)$$

Where $x_{dk}$ is the $d_{th}$ feature of the $k_{th}$ pattern and $\mu_d$ is the respective mean value of the attribute.

3. This method uses of the equation shown in Eq. (5) (Max-Min method)

$$\frac{\max\ diff + \min\ diff}{2} \quad (5)$$

## 4. KFLANN Algorithm

### 4.1 KFLANN Architecture

The basic architecture of KFLANN is delineated in Figure 1. It consists of two layers, similar to the basic Kohonen network model [2] and the ART1 model [4]. Connecting the output layer and the input layer is a set of weight vectors. Each output node has weights connected to each element of the input vector. The weight assignment system is unlike that of the Kohonen network or ART1. If a novel pattern is to be stored, it performs a direct copy of the input vector into the weight vector. No other forms of equation are needed for evaluating the weight vectors. The output layer can be viewed as a single dimensional layer which grows dynamically as more novel patterns are encountered. As these novel patterns are encountered, the output system will allocate memory space for the pattern. This implies that the design stages of the network would not require a detailed plan of the output layer, but only a correct configuration of the input layer. The output layer will determine its own size, as the system is trained. Classification results are eventually obtained directly from the winning output node.

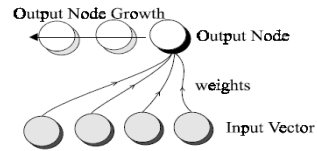

### 4.2 Algorithm

Step 1   Initialize the network parameters.

Step 2   Present the pattern to the input layer. If there is no output node, GOTO step 6

Step 3   Determine all possible matches output node using Eq (6).

$$\frac{\sum_{i=0}^{n}\left[\delta_i^2 - (w_{ij} - x_i)^2\right]}{n} \geq \rho \quad (6)$$

Step 4    Determine the winner from all matches output nodes using Eq.(7)

$$\text{Winner} = \min\left[\sum_{i=0}^{n}(w_{ij} - x_i)^2\right] \quad (7)$$

Step 5    Match node is found. Assign the pattern to the match output node. GOTO Step 2

Step 6    Create new output node. Perform direct mapping of the input vector into weight vectors.

Step 7    If complete a single epoch compute clusters centroid. If centroid points of all clusters unchanged Terminate

Else

 GO TO Step 2.

Step 8    Find closest pattern to the centroid and re-shuffle it to the top of the dataset list, GOTO Step 2.

*Note*:   ρ is the Vigilance Value,    $\delta_i$    is the tolerance for $i_{th}$ feature of the input space , $W_{ij}$ used to denote the weight connection of $j^{th}$ output to the $i^{th}$ input, $X_i$ represent the $i^{th}$ feature.

### 4.3 Tolerance Tuning

Using the domain knowledge it is possible to know number of clusters in the given data set. After completing a single epoch, if desired number of clusters are not formed then KFLANN requires tolerance tuning .It is given as follows:

*i.*   Initialize              $\delta_i = (\delta_{imax} + \delta_{imin})/2$      (8)
ii.   While number of clusters formed is not appropriate
iii.  run KFLANN algorithm without step 7 based on current $\delta$, values
iv.   if number of clusters is less than expected
              $\delta_i = (\delta_i + \delta_{imin}) / 2$           (9)
v.    else if number of clusters is more than expected
        $\delta_i = (\delta_i + \delta_{imax}) / 2$              (10)
vi.   end while

$\delta_i$ - Tolerance value for attribute i. $\delta_{imin}$ - Minimum difference in attribute values for attribute *i*. This is the difference between the smallest and the second smallest values of the attribute. $\delta_{imax}$ - **Maximum** difference in attribute values for attribute *i*. This is the difference between the smallest and largest values of the attribute.

## 5. Experimental Results

### 5.1 Data Sets

#### 5.1.1 Artificial Data Sets Used

The data sets that are used to test the KFLANN algorithm are obtained from the site http://kdd.ics.uci.edu/)

Table 1: Description of Artificial Data Sets

| Data Set | # of patterns | # of features | # of clusters |
|---|---|---|---|
| Iris | 150 | 4 | 3(class 1-50,class 2-50,class 3-50) |
| Wine | 178 | 13 | 3(class 1-59,class 2-70,class 3-49) |
| Glass | 214 | 9 | 7(class 1-70,class 2-76,class 3-17,class 4-0,class 5-13,class 6-9,class 7-29) |
| Haberman | 306 | 3 | 2(class 1-227,class 2-79) |
| New Thyroid | 215 | 5 | 3(class 1-150,class 2-35,class 3-30) |
| Image Segmentation | 210 | 19 | 7(30 per class) |
| Pima-Indian diabetes | 768 | 8 | 2(class 1-500,class 2-268) |
| Ionosphere | 351 | 34 | 2(class 1-225,class 2-126) |

#### 5.1.2 Synthetic Data Sets Used

Table 2: Description of Synthetic Data Sets

| Data Set | # of patterns | #of features | # of clusters |
|---|---|---|---|
| Synthetic Data Set 1(Well separated) | 1000 | 2 | 2(class 1-500,class 2-500) |
| Synthetic Data Set 2(Half separated) | 1000 | 2 | 2(class 1-500,class 2-500) |
| Synthetic Data Set 3(Not separated) | 1000 | 2 | 2(class 1-500,class 2-500) |
| Synthetic Data Set 4 | 500 | 8 | 3(class 1-250,class 2-150,class 3-100 ) |
| Synthetic Data Set 5 | 400 | 8 | 3(class 1-150,class 2-150,class 3-100 ) |
| Synthetic Data Set 6 | 350 | 8 | 3(class 1-100,class 2-150,class 3-100 ) |

### 5.2 Setting the Vigilance Parameter (ρ) and Tolerance Parameter (δ) is fixed for each attribute.

The following plots shows the number of clusters formed with the given vigilance value. The groups formed in the data grow with the increase in the vigilance value.

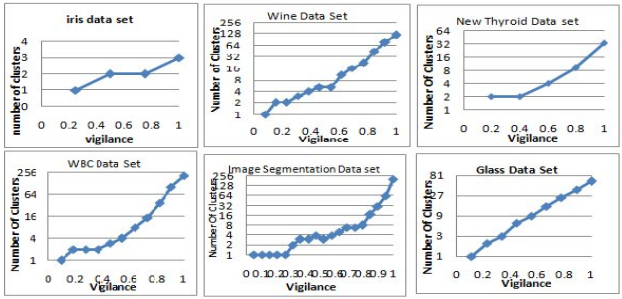
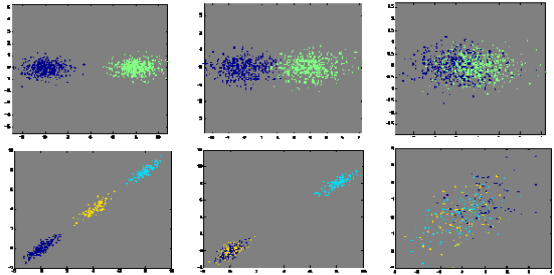

Fig. 2 Plots of Synthetic Data Sets

### 5.3 Non-normalized data

The Experiments are conducted with non-normalized input data with the given neural network input parameters namely vigilance and tolerance. The following table shows the misclassification rate when Eq. (5) was used for computing tolerance setting for each attribute.

Table 3: Results with non-normalized data and tolerance with Eq.5

| data set | Vigilance (ρ) | #of clusters | With Eq. (5) Error Rate (%) |
|---|---|---|---|
| Iris | 1 | 3 | 7.3333 |
| New Thyroid | 1 | 4(3 is the actual value, (tolerance tuning is required)) | 20.4651 |
| New Thyroid | 0.8 | 2(3 is the actual value, (tolerance tuning is required)) | 24.6512 |
| Ionosphere | 0.5294 | 2 | 29.6296 |
| Pima Indian diabetes | 0.6250 | 2 | 33.4635 |
| Wine | 0.7692 | 3 | 35.3933 |
| Glass | 0.8889 | 7 | 54.6729 |
| Haberman | 0.3333 | 6(2 is the actual value, tolerance tuning is required) | 70.2614 |
| Image Segmentation | 0.8947,0.9474 | 4,8(7 is the actual value, tolerance tuning is required) | 79.0476, 64.4286 |

The following table shows the misclassification rate when (standard deviation i.e. Eq. (4) was used for computing tolerance setting for each attribute.

Table 4: Results with non-normalized data and tolerance With Eq.4

| data set | Vigilance (ρ) | #of clusters | With Eq. (4) Error Rate (%) |
|---|---|---|---|
| Iris | 0.5 | 3 | 10 |
| New Thyroid | 0.4 | 2(3 is the actual value, tolerance tuning is required) | 30.2326 |
| Ionosphere | 0.2647 | 2 | 31.3390 |
| Pima Indian diabetes | 0.2500 | 2 | 34.2448 |
| wine | 0.3077 | 3 | 38.2022 |
| New Thyroid | 0.6 | 4(3 is the actual value ,tolerance tuning is required) | 45.5814 |
| Haberman | 0.3333 | 2 | 48.6928 |
| Image Segmentation | 0.6842,0.7368 | 7 | 64.2857,64.7619 |
| Glass | 0.4444 | 7 | 67.2897 |

The above results show that the error rate was low when Eq. (5) was used for computing tolerance than Eq. (4) was used .KFLANN performs well only for iris data. For New Thyroid data set and Image Segmentation, actual number of clusters are not formed with the given vigilance value, so tolerance tuning was required. So Normalization was required in order to classify samples with lower misclassification.

### 5.4 Normalized data

#### 5.4.1 Z-score normalization

Table 5: Results with Z-Score and tolerance with Eq.5

| Data Set | Vigilance (ρ) | #of Clusters | With Eq. (5) Error Rate (%) |
|---|---|---|---|
| Iris | 0.75 | 3 | 5.3333 |
| New Thyroid | 0.8 | 4(actual value is 3) | 16.7442 |
| Haberman | 0.6667 | 2 | 25.4092 |
| wine | 0.9231 | 3 | 32.5843 |
| Ionosphere | 0.8235 | 2 | 42.4501 |
| Pima Indian diabetes | 0.75 | 2 | 48.4375 |
| Glass | 0.8889 | 7 | 52.8037 |
| Image Segmentation | 0.9,1 | 6 ,10(actual value is 7) | 80.25, 82.30 |

Table 6: Results with Z-Score and tolerance with Eq.4

| data set | Vigilance (ρ) | #of clusters | With Eq. (4) Error Rate (%) |
|---|---|---|---|
| Iris | 0.5 | 3 | 5.3333 |
| New Thyroid | 0.2 | 3 | 13.4884 |
| Ionosphere | 0.3235 | 2 | 40.1709 |
| wine | 0.4615 | 3 | 43.2584 |
| Pima Indian diabetes | 0.1250 | 2 | 48.3073 |
| Haberman | 0.3333 | 2 | 53.2680 |
| Glass | 0.4444 | 7 | 56.0748 |
| Image Segmentation | 0.6842 | 6 (actual value is 7,tolerance tuning is required) | 87.6190 |

### 5.4.2 Max-Min Normalization

Table 7: Results with Max-Min and tolerance with Eq.5

| Data Set | Vigilance (ρ) | #of clusters | With Eq.(5)Error Rate (%) |
|---|---|---|---|
| Pima Indian diabetes | 0.6250 | 2 | 0 |
| Iris | 1 | 3 | 10 |
| wine | 0.7692 | 3 | 14.6067 |
| New Thyroid | 0.8,1 | 2,4(actual value is 3,tolerance tuning is required) | 21.8605,12.5581 |
| Ionosphere | 0.5294 | 2 | 29.6296 |
| Glass | 0.8889 | 7 | 49.5327 |
| Image Segmentation | 0.9474 | 8 (tolerance tuning is required) | 59.5238 |
| Haberman | 0.6667,1 | 3,6(actual value is 2,tolerance tuning is required) | 60.7843,80.7190 |

Table 8: Results with Max-Min and tolerance with Eq.4

| data set | Vigilance (ρ) | #of clusters | With Eq.(4) Error Rate (%) |
|---|---|---|---|
| Pima Indian diabetes | 0.2500 | 2 | 0.1302 |
| Iris | 0.5 | 3 | 12 |
| New Thyroid | 0.4,1 | 2,4(tolerance tuning is required) | 30.2326,12.5581 |
| Ionosphere | 0.2647 | 2 | 31.3390 |
| Haberman | 0.3333 | 2 | 54.5752 |
| Image Segmentation | 0.7368 | 7 | 57.1429 |
| Wine | 0.3077 | 3 | 64.6067 |
| Glass | 0.4444 | 7 | 65.8879 |

With the above results, it was observed that accuracy is high with normalized data. KFLANN performs well for Pima Indian Diabetes data set where the error rate is zero. Using Z-score normalization for the Iris data set the error rate was same when either of the method (stated above) was used for computing tolerance. It was found that Z-score Normalization was well suited for Haberman survival data and Max-Min normalization was well suited for Wine, Ionosphere and Glass (tolerance setting with Max-Min) data sets as the error rate was low when compared with Z-Score Normalization.

### 5.5 Why tolerance tuning?

From the above results, it was found that desired number of clusters is not formed with maximum vigilance value (1) and also with the next maximum value. So there is no chance of getting stated number of clusters because if the vigilance value is reduced number of clusters formed is less than the previous value. Getting number of clusters stated in the domain knowledge is possible only with tolerance tuning when max-min method was used for setting tolerance parameter for the attributes. For some of the artificial data sets tolerance tuning is required as expected number of clusters is not formed. For example consider New Thyroid Data set .The number of clusters are 3.but with the given vigilance values 2 and 4 are formed.

### 5.5.1 Synthetic Data Sets

On all the Synthetic data sets, KFLANN requires tolerance tuning. The misclassification rate was as follows:

Table 9: Results with Synthetic Data sets

| Data Set | Vigilance (ρ) | #of clusters | Eq. (4) or Eq. (5) Error Rate (%) |
|---|---|---|---|
| Synthetic Data Set 1 | 1 | 2 | 0 |
| Synthetic Data Set 2 | 1 | 2 | 1.8 |
| Synthetic Data Set 3 | 1 | 2 | 29.2 |
| Synthetic Data Set 4 | 1 | 3 | 0 |
| Synthetic Data Set 5 | 1 | 3 | 0 |
| Synthetic Data Set 6 | 1 | 3 | 54.2857 |

## 6. Conclusions

This paper was intended to observe the behavior of a KFLANN algorithm used for unsupervised learning. The above results show that the performance of the algorithm and the quality of the clustering was based on the input data presented (normalized or non-normalized). The result of the KFLANN was varied between non-normalized and normalized data and was not same for all types of data normalization. For a given input data, one of the normalization method stated above was well suited for a given input because the efficiency of the KFLANN was improved. So normalization is one of the most important factors in clustering. The operation of the KFLANN also depends on above mentioned two parameters namely vigilance and tolerance. It was observed that the misclassification rate is low when Max-Min formula was used for tolerance and also independent of normalization method. For all synthetic data sets the misclassification rate is same with the vigilance setting and with different formulas used for tolerance setting. A comparative analysis showed that the clustering results depended on the normalization method used and the noisiness of the data. In particular, the selection of the vigilance and tolerance setting values for the KFLANN algorithm was sensitive to the normalization method used for datasets with large variations across samples. The fine behavior obtained with neural network for the clustering problems studied here can be used with other difficult optimization problems.

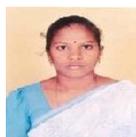
**[1]Suneetha Chittineni**, Associate Professor in the department of Computer Applications, R.V.R.& J.C college of Engineering,Chowdavarm,Guntur. She has 12 years of teaching experience. Currently she is pursuing Ph.D. from Acharya Nagarjuna University, Guntur. Her research interests include Artificial Intelligence, Machine learning, Pattern Recognition.

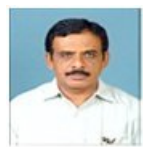
**[2]Dr B. Raveendra Babu** obtained his Masters in Computer Science and Engineering from Anna University, Chennai. He received his Ph.D. in Applied Mathematics at S.V University, Tirupati. He is currently leading a Team as Director (Operations), M/s. Delta Technologies (P) Ltd., Madhapur, Hyderabad. He has 26 years of teaching experience. He has more than 25 International & National publications to his credit. His research areas of interest include VLDB, Image Processing, Pattern Analysis and Information Security.